\documentclass[letterpaper, 10 pt, conference]{ieeeconf}  

\IEEEoverridecommandlockouts                              
                                                          
\usepackage{color}
\usepackage{multirow}
\usepackage{booktabs} 
\usepackage{url}
\usepackage{subfigure}
\usepackage{cite}

\usepackage{subfloat}
\usepackage{subfigure}
\usepackage[pdftex]{graphicx}
\graphicspath{{figures/}}

\usepackage{enumerate}
\usepackage{amssymb}
\usepackage{leftidx}
\usepackage{amsfonts}
\usepackage{amsmath}
\usepackage{bm}
\usepackage{booktabs}
\usepackage{setspace}
\usepackage{makecell}
\usepackage[colorlinks,linkcolor=blue]{hyperref}
\usepackage[ruled,linesnumbered]{algorithm2e}
\usepackage{mathrsfs}

\usepackage{colortbl}
\usepackage[table]{xcolor}

\usepackage{pbox}
\usepackage{bm} 
\usepackage{multirow}
\usepackage{diagbox}
\usepackage{ulem}

\usepackage{booktabs}
\usepackage{makecell}
\usepackage{bbding}
\usepackage{utfsym}
\usepackage{threeparttable}
\usepackage{soul}

\begin{document}
\title{Multi-object Detection, Tracking and Prediction in Rugged Dynamic Environments}

\author{Shixing Huang, Zhihao Wang, Junyuan Ouyang, Haoyao Chen*
	\thanks{This work was supported in part by the National Natural Science Foundation of China under Grant U21A20119, and the Shenzhen Science and Innovation Committee under JCYJ20200109113412326. (Corresponding author: Haoyao Chen.)}
	\thanks{S.X. Huang, Z.H. Wang, J.Y. Ouyang, and H.Y. Chen* are with the School of Mechanical Engineering and Automation, Harbin Institute of Technology Shenzhen, P.R. China, e-mail: hychen5@hit.edu.cn.}
}
\maketitle
\begin{abstract}
  Multi-object tracking (MOT) has important applications in monitoring, logistics, and other fields. This paper develops a real-time multi-object tracking and prediction system in rugged environments. A 3D object detection algorithm based on Lidar-camera fusion is designed to detect the target objects. Based on the Hungarian algorithm, this paper designs a 3D multi-object tracking algorithm with an adaptive threshold to realize the stable matching and tracking of the objects. We combine Memory Augmented Neural Networks (MANN) and Kalman filter to achieve 3D trajectory prediction on rugged terrains. Besides, we realize a new dynamic SLAM by using the results of multi-object tracking to remove dynamic points for better SLAM performance and static map. To verify the effectiveness of the proposed multi-object tracking and prediction system, several simulations and physical experiments are conducted. The results show that the proposed system can track dynamic objects and provide future trajectory and a more clean static map in real-time.
\end{abstract} 
\IEEEpeerreviewmaketitle

  
\section{Introduction}
Multi-object tracking refers to detecting and storing information about target objects of interest in the sensor's field of view (FOV). In recent years, multi-object tracking has been widely used in robotics applications such as aerial videography \cite{nageli2017real}, monitoring security, and logistics field. 
Camera and Lidar are commonly used in object tracking. The camera provides images with rich color and texture information, while Lidar provides point clouds with accurate depth information. Thus, many recent works \cite{vora2020pointpainting,CharlesRQi2017FrustumPF} focused on fusing Lidar-camera to obtain abundant and versatile information on the appearance and motion of objects for robust and accurate tracking. 
As images and 3D Lidar point cloud are data of different modalities, the methods to achieve multimodal data fusion are divided into data-level fusion, feature-level fusion \cite{vora2020pointpainting}, and result-level fusion \cite{CharlesRQi2017FrustumPF}.

Most object tracking methods \cite{wang2022deepfusionmot,XinshuoWeng2020AB3DMOTAB,ErkanBaser2019FANTrack3M} are devised under a tracking-by-detection framework. The framework detects the objects and then tracks the objects by data association. The data association will judge whether the distance between the detected and predicted position exceeds a preset threshold to determine whether the matching is successful.
However, using the preset fixed threshold leads to a low success rate of matching for the targets with non-uniform and change-of-direction motion.

\begin{figure}[!t]
  \centering
  \includegraphics[width=3.35in]{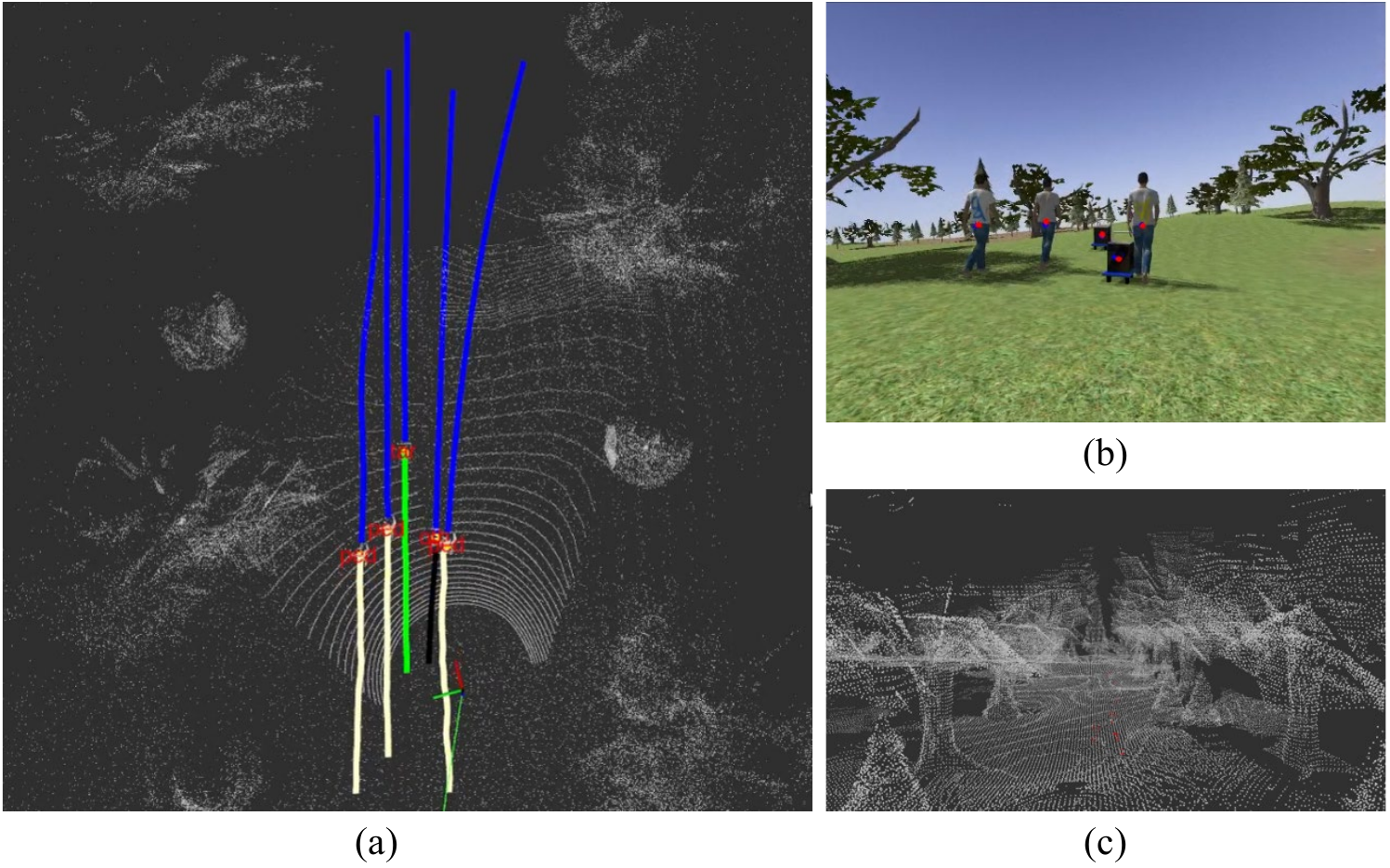}
  \caption{Experimental results of the proposed method. (a) Multi-object detection, tracking, and prediction results. The white, green, and black trajectories are the history trajectories, the blue ones are the predicted trajectories, and the red fonts are the object categories. (b) Camera view. (c) Static map after removing dynamic points. }
  \label{fig:cover}
\end{figure}

Object tracking generates historical trajectories of the dynamic object. While predicting future states of the dynamic object is also essential, which benefits object tracking and ego vehicle's motion decision \cite{8793868}. The traditional trajectory prediction methods, such as least squares fitting and Kalman filter \cite{prevost2007extended}, can correctly predict future states in a short time horizon (such as 1 second). 
But these traditional methods do not perform well to predict a longer time horizon. Many researchers \cite{AmirSadeghian2019SoPhieAA,FrancescoMarchetti2020MANTRAMA,RohanChandra2019TraPHicTP} have used the advantages of deep learning in feature extraction to solve the prediction problem. 
However, most existing methods of trajectory prediction based on deep learning \cite{AmirSadeghian2019SoPhieAA,FrancescoMarchetti2020MANTRAMA,RohanChandra2019TraPHicTP} only consider the objects moving in 2D space. These methods may lead to prediction errors when objects move on rough terrain in outdoor environments.

Our work considers a robot following the target objects in real-time and uses the existing SLAM scheme for locating the robot and the objects. The dynamic objects, including the target objects and other interfering objects, will affect the performance of SLAM. 
Besides, mapping the environment where the robot has traveled can be used for data recording or relocalization. This situation requires a clean and static map to remove the influence of dynamic objects.
Thus, dynamic SLAM is widely researched to remove dynamic objects and points.
Existing works, including visibility-based \cite{GiseopKim2020RemoveTR}, {ray-casting based} \cite{lim2021erasor}, detection-based \cite{9830851}, and segmentation-based \cite{chen2019sumapp} methods, detect the dynamic objects in the current frame and then filter out the points around the dynamic objects. These methods perform well by using single frame data to deal with dynamic SLAM in plane terrain or autonomous driving scenario.
But dynamic object detection based on single frame data often has missed detection and false detection, especially in complex environments \cite{9939009}. These situations cause the dynamic objects to leave unwanted traces on the map.

This paper aims to develop a multi-object detection, tracking, and prediction system in rugged dynamic environments by resolving the abovementioned problems. As shown in Fig. \ref{fig:cover}, the system provides the history and predicted trajectories of dynamic objects and creates a high-quality static map. The main contributions include the following three points:
\begin{itemize}
    \item {To solve the problem of matching errors under non-uniform and change-of-direction motion in the traditional matching method, a threshold-adaptive 3D multi-object tracking algorithm is proposed to improve the accuracy of multi-object tracking.}

    \item {A new trajectory prediction method is proposed to decrease the prediction error when objects move on rugged terrains. Moreover, the method combines Memory Augmented Neural Networks and Kalman filter to improve the robustness of trajectory prediction. }

    \item {A real-time dynamic points removal method based on the tracking and prediction system is developed to enhance SLAM performance and obtain a better static map. }
\end{itemize}

\section{METHODOLOGY}

\begin{figure}[htbp]
    \centering
    \includegraphics[width=3.4in]{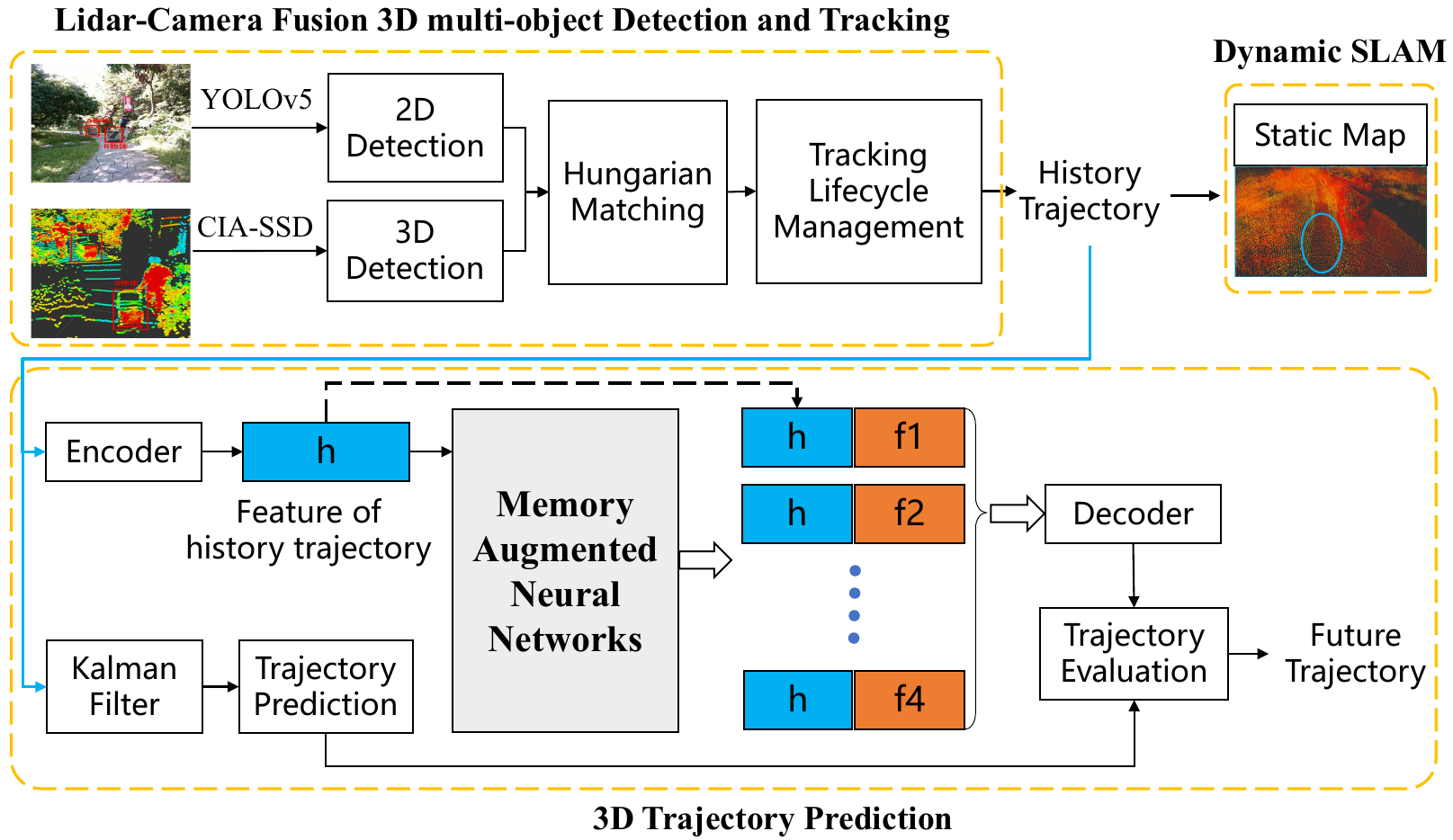}
    \caption{System Framework}
    \label{fig:pipeline}
\end{figure}

The proposed multi-object detection, tracking, and prediction system, as shown in Fig. \ref{fig:pipeline}, consists of three parts: (1) multi-object detection and tracking based on Lidar-camera fusion; (2) 3D trajectory prediction; (3) dynamic SLAM.

\subsection{Multi-object detection and matching}
\begin{figure}[htpb]
	\centering
	\includegraphics[width = 3.4in]{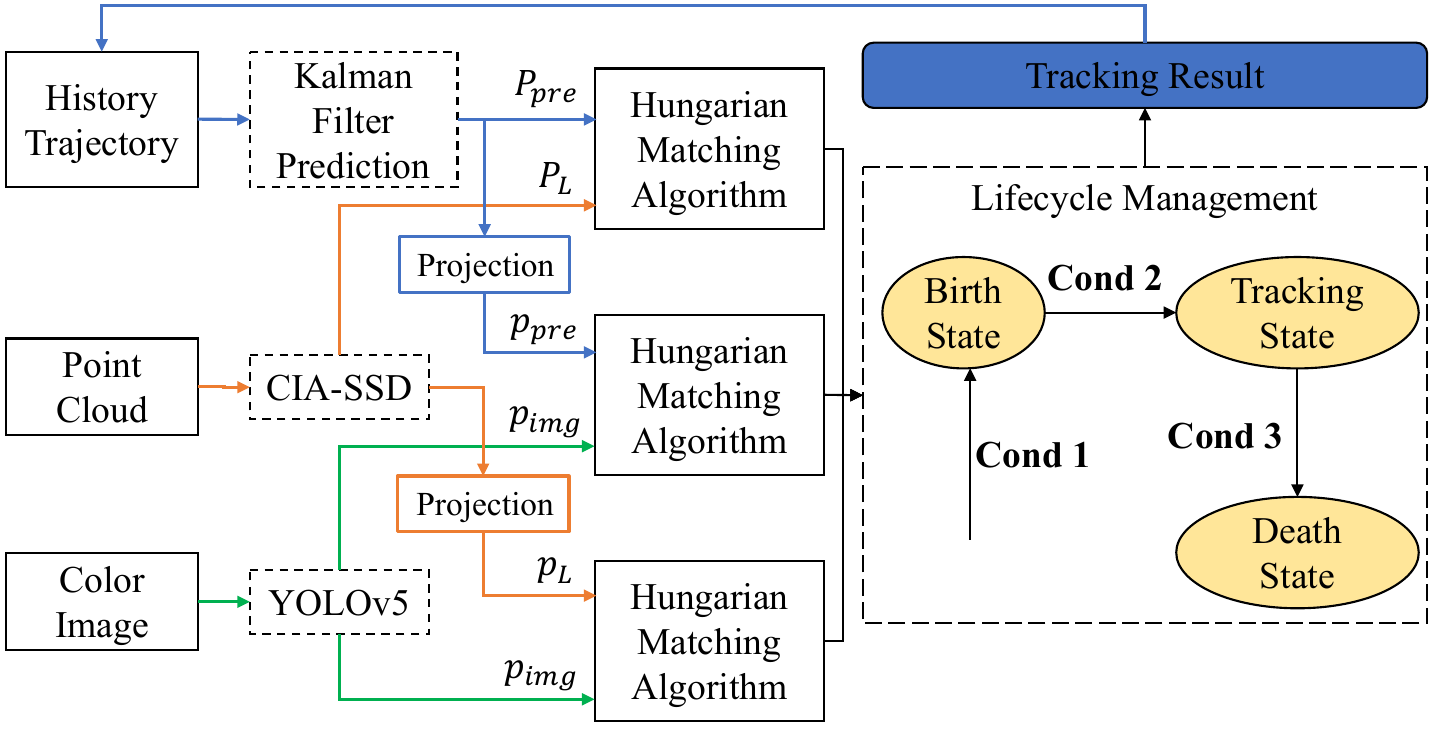}
	\caption{Matching strategy and tracking instance life management framework}
	\label{fig:Target_matching}
\end{figure}

A target object $j$ in one frame of sensor data is an example to show the object detection and matching process. 
We use CIA-SSD \cite{WuZheng2021CIASSDCI}, which balances computational efficiency and detection accuracy, to complete the 3D object detection by LiDAR point cloud. 
The 3D detection result $P_L^j$ represents the center coordinate of the 3D bounding box of the object in the LiDAR coordinate system, and its projection to the image coordinate system is $p_L^j$. 
The YOLOv5 algorithm \cite{githubYolov5} is used to detect the center coordinate of the 2D bounding box of the object in the image, which is given as $p_{img}^j$. Note that the detection method used in this work is not limited to CIA-SSD or YOLOv5, other similar algorithms that can provide detection results also work. 
If the object is detected by both the LiDAR and camera, a \textbf{tracking instance} of the object is established. 
The position of the object $P_{pre}^j$ is predicted by using Kalman filter \cite{prevost2007extended}. The Kalman filter takes the kinematics model of the tracking instance as the prediction model and takes the detected position of the object by point cloud and image fusion as the observation. The projection of $P_{pre}^j$ in the 2D image is $p_{pre}^j$. 
To complete the matching of the above detection and prediction results, Hungarian algorithm \cite{kuhn2005hungarian} is used for data association by calculating Euclidean distance between $P_{pre}^j$ and $P_L^j$, $p_{pre}^j$ and $p_{img}^j$, and $p_L^j$ and $p_{img}^j$ to construct a cost matrix. 
Traditional matching algorithms determine whether two positions are matched by judging whether the Euclidean distance exceeds a preset threshold. The preset fixed threshold leads to tracking errors in some cases, like non-uniform motion. 
Therefore, an adaptive threshold solution is proposed by using the object's motion information to set a dynamic threshold to prevent incorrect matches. By taking the matching of $P_{pre}^j$ and $P_L^j$ as an example, the adaptive threshold is formulated as
\begin{equation}
  \theta_{th}=a \cdot \sqrt{v_x^2+v_y^2}+b
\end{equation}
where $\theta_{th}$ denotes the dynamic threshold for matching $P_{pre}^j$ and $P_L^j$; $a$, $b$ are constant coefficients; $v_x$, $v_y$ are the velocity in $x$, $y$ direction estimated by Kalman filter.

\subsection{Tracking lifecycle management}
Object tracking not only completes the matching of new data in each frame with historical trajectories to update the instance's state but also needs to manage the instance's state. 
Specifically, there are mainly three states in tracking process: (1) the instance may leave the sensor's FOV or sensing range permanently, (2) a new instance will suddenly enter the sensor's FOV, and (3) the existing instance remains in the sensor's FOV. Therefore, these states need to be recorded and managed. The process of management is called lifecycle management.
To solve the lifecycle management of multi-object tracking with multi-sensor, we propose a strategy based on the tracking results in multi-frame data to judge the states of instances. 
The states include birth state (a new instance enters the sensor's FOV), tracking state (the existing instance remains within the sensor's FOV), and death state (the existing instance leaves the sensor's FOV). The detection logic of the state transition is shown as TABLE \ref{table:lifecycle} and the lifecycle management part in Fig. \ref{fig:Target_matching}. Specifically, the below conditions decide the state transition.
\textbf{Condition 1}: When the detection result in the 2D image $p_{img}^j$ matches the projection of the corresponding 3D LiDAR detection in the image $p_L^j$, while $p_{img}^j$ cannot match the predicted value of all tracked instances $p_{pre}$, it means that a new object has entered the detection range, and an instance will be created for this object. The instance is in the birth state.
\textbf{Condition 2}: When the detection result $p_{img}^j$ matches $p_L^j$, $p_L^j$ matches $p_{pre}^j$, and $p_{img}^j$ matches $p_{pre}^j$ or when $p_L^j$ matches $p_{pre}^j$ but $p_{img}^j$ does not match $p_{pre}^j$, $P_L^j$ is used as the measurement of corresponding instance. If the measurement of an instance is detected in three consecutive frames of sensor data, the state of the instance translates from the birth state to the tracking state.
\textbf{Condition 3}: When the predicted value $p_{pre}^j$ of an instance cannot find any matching object both on the 2D image data and the projection of 3D LiDAR data, the predicted 3D position $P_{pre}^j$ is used as the current position of the instance temporarily. If $p_{pre}^j$ does not find a matching object in five consecutive frames of sensor data, we consider that the object $j$ has left the sensor's FOV and then delete the instance. The instance enters the death state.

\begin{table}[]
  \caption{the logic table of state transition.}
  \centering
  \begin{threeparttable} 
  \begin{tabular}{|c|c|c|c|c|c|}
    \hline
                                                                &                             & $p_{img}^j$                                & $p_{L}^j$                         & $p_{pre}^j$                       & state                               \\ \hline
    \textbf{Cond 1}                                             & $p_{img}^j$                         &                                    & \usym{2714}                         & \usym{2718}                         & \textbf{Birth}                      \\ \hline
    \cellcolor[HTML]{EFEFEF}                                    & \cellcolor[HTML]{EFEFEF}$p_{img}^j$ & \cellcolor[HTML]{EFEFEF}           & \cellcolor[HTML]{EFEFEF}\usym{2714} & \cellcolor[HTML]{EFEFEF}\usym{2714} &                                     \\ \cline{2-5}
    \multirow{-2}{*}{\cellcolor[HTML]{EFEFEF}\textbf{Cond 2.1}} & \cellcolor[HTML]{EFEFEF}$p_{L}^j$   & \cellcolor[HTML]{EFEFEF}\textbf{\usym{2714}} & \cellcolor[HTML]{EFEFEF}  & \cellcolor[HTML]{EFEFEF}\usym{2714} &                                     \\ \cline{1-5}
    \cellcolor[HTML]{C0C0C0}\textbf{Cond 2.2}                   & \cellcolor[HTML]{C0C0C0}$p_{L}^j$   & \cellcolor[HTML]{C0C0C0}\usym{2718}          & \cellcolor[HTML]{C0C0C0}  & \cellcolor[HTML]{C0C0C0}\usym{2714} & \multirow{-3}{*}{\textbf{Tracking}} \\ \hline
    \textbf{Cond 3}                                             & no match                    & \textbf{\usym{2718}}                         & \usym{2718}                         & \usym{2718}                         & \textbf{Death}                      \\ \hline
    \end{tabular}
    \begin{tablenotes}    
      \footnotesize               
      \item[1] \usym{2714} denote the data of corresponding row and column match successfully, \usym{2718} denote the data cannot match.          
    \end{tablenotes}            
    \end{threeparttable} 
    \label{table:lifecycle}
\end{table}


\subsection{Trajectory prediction}
Deep neural networks can extract motion information as features from the trajectory. Memory Augmented Neural Networks (MANN) \cite{FrancescoMarchetti2020MANTRAMA} learns features of history and future trajectory and exploits an associative external memory to store and retrieve such features. 
Trajectory prediction is then performed by decoding futures conditioned with the observed history states.
Our trajectory prediction algorithm is designed based on MANTRA \cite{FrancescoMarchetti2020MANTRAMA}, which includes an encoder network to extract trajectory features, a decoder network to recover future trajectory, a memory controller, and memory for storing features. 
The structure of the encoder network is composed of 1D Convolution Neural Network and Gated Recurrent Units. The decoder network consists of Gated Recurrent Units and fully connected layers. 
The encoder network is divided into a history trajectory encoder network and a future trajectory encoder network, which have the same network structure but different learning data. 
The history trajectory feature $hi$ and the future trajectory feature $fi$ are obtained by the history and future trajectory encoders, respectively.
Then $hi$ and $fi$ are spliced and can be put into the decoder network to recover the future trajectory of an object.
We train the encoder and decoder networks by recording data when the robot and targets move on 3D rugged terrains. And the history and future trajectories used in the training process are in 3D space. 

After training the encoder and decoder networks, the history trajectory encoder obtains the history trajectory feature $h$. The cosine distance is used to find five history trajectory feature vectors most similar to $h$ in the history trajectory feature memory, and further find the corresponding five future trajectories.
We aim to select the optimal future trajectory by evaluating these five trajectories. 
Kalman filter performed well in the prediction of short-term trajectory. Thus, we use the trajectory predicted by Kalman filter (1-second trajectory) as an indicator to select the optimal predicted trajectory. 
The trajectory points $\hat{P}_{F\_kf}$ in the next 1 second are estimated and then used to calculate the error of the 1-second trajectory points to the candidate future trajectories. The future trajectory with the smallest error is selected as the optimal trajectory prediction.

In addition, MANTRA needs the history trajectory of objects in a period to complete the prediction. If the number of history trajectory points of the objects cannot meet the requirements, such as the initial stage of the tracking process, MANTRA cannot extract useful features. At this time, Kalman filter is used to complete the trajectory prediction based on the preset motion model and current observations. Thus, the combination of MANTRA and Kalman filter enhances the robustness of trajectory prediction.

\subsection{Dynamic SLAM}
LIO-SAM \cite{TixiaoShan2020LIOSAMTL} is adopted as the SLAM scheme to estimate the robot's pose. But the dynamic objects affect the performance of SLAM.
Traditional dynamic SLAM removes the points of dynamic objects depending on the 3D object detection result. But the 3D object detection algorithm may miss/falsely detect a moving object with a single data frame, which leads to the dynamic points not being removed. 
To solve this problem, we remove the dynamic points based on the object tracking result rather than the detection with a single data frame.
Specifically, we use the last position of the tracked history trajectory as the position of the dynamic object to remove dynamic points. In this way, even if the detection of the current data frame is missed or false, the result inferred from tracked history trajectory can also provide information on dynamic object to remove dynamic points. 
After removing the dynamic points, the incremental map is constructed using ikd-Tree \cite{YixiCai2021ikdTreeAI} to establish the static map.

In addition, when the points of dynamic objects are not removed due to the wrong tracking result, the residual error from the dynamic points to the corresponding plane is relatively large. 
Therefore, we calculate the error value by using \eqref{equ:relative_error}. If the value is greater than the preset threshold, the point is a dynamic point or noise point, and it is removed. 
\begin{equation}
	s = \frac{r_{plane}}{\sqrt[]{\Vert P_i \Vert}}
	\label{equ:relative_error}
\end{equation}
where $P_i$ represents the coordinates of the $i$th feature point of the point cloud frame; $r_{plane}$ represents the residual of the point cloud point $P_i$ to the nearest plane of the local map.

\section{SIMULATIONS AND EXPERIMENTS}
\begin{figure}[htpb]
	\centering
	\includegraphics[width = 3.4in]{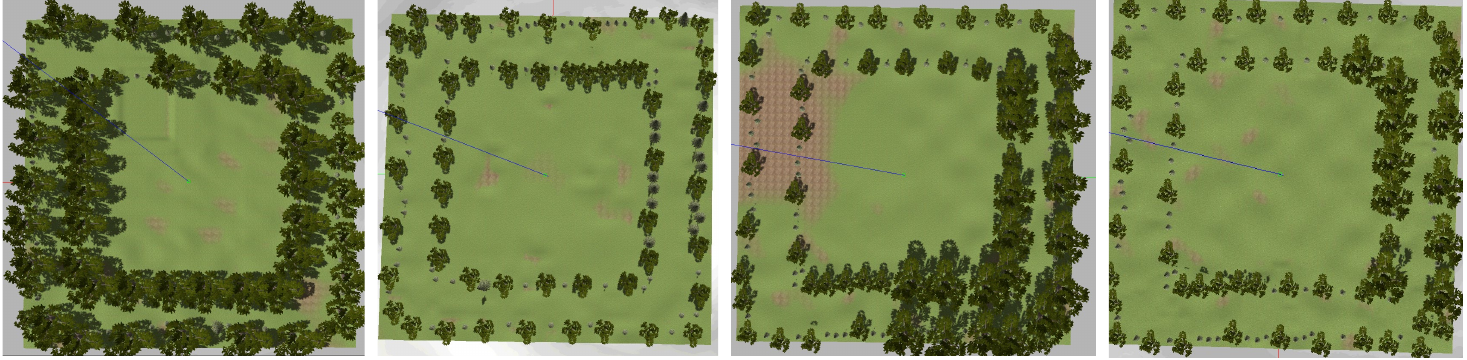}
	\caption{Simulation environments}
	\label{fig:training_environment_simulation}
\end{figure}

\begin{figure}[htpb]
	\centering
	\includegraphics[width = 3.4in]{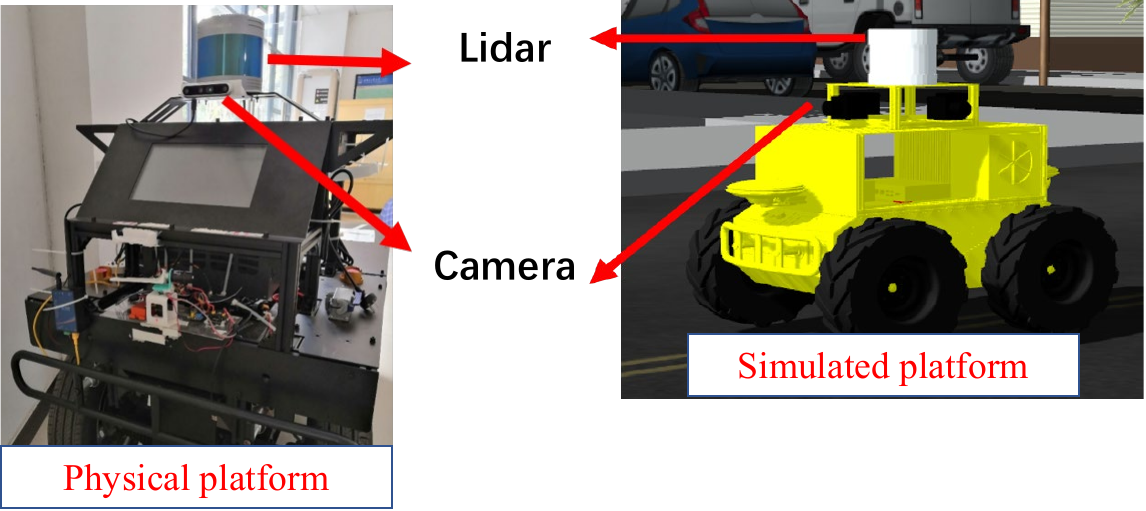}
	\caption{Robot platform}
	\label{fig:the robot physical platform}
\end{figure}

To verify the effectiveness of the proposed methods, we carried out several simulations and physical experiments. 
As shown in Fig. \ref{fig:training_environment_simulation}, we use Gazebo simulator to build eight rough forest scenarios, which contain static objects (trees), dynamic objects (pedestrians and moving objects), vegetation, and terrain with varying degrees of ruggedness. 
The robot built in the Gazebo, as shown in Fig. \ref{fig:the robot physical platform}, has the RS-Ruby Lite (80-line Lidar), a monocular camera with a resolution of 640$\times$480, and an IMU inside the robot. 
The point cloud, images, and IMU data are recorded and made into training and testing datasets. 
The five pieces of datasets of sim03-{sim07} are used as training datasets for CIA-SSD and YOLOv5. The three pieces of data sim00-{sim02} are used as testing datasets, with a total of 3655 frames of data.

\begin{figure}[htpb]
	\centering
	\includegraphics[width = 3.4in]{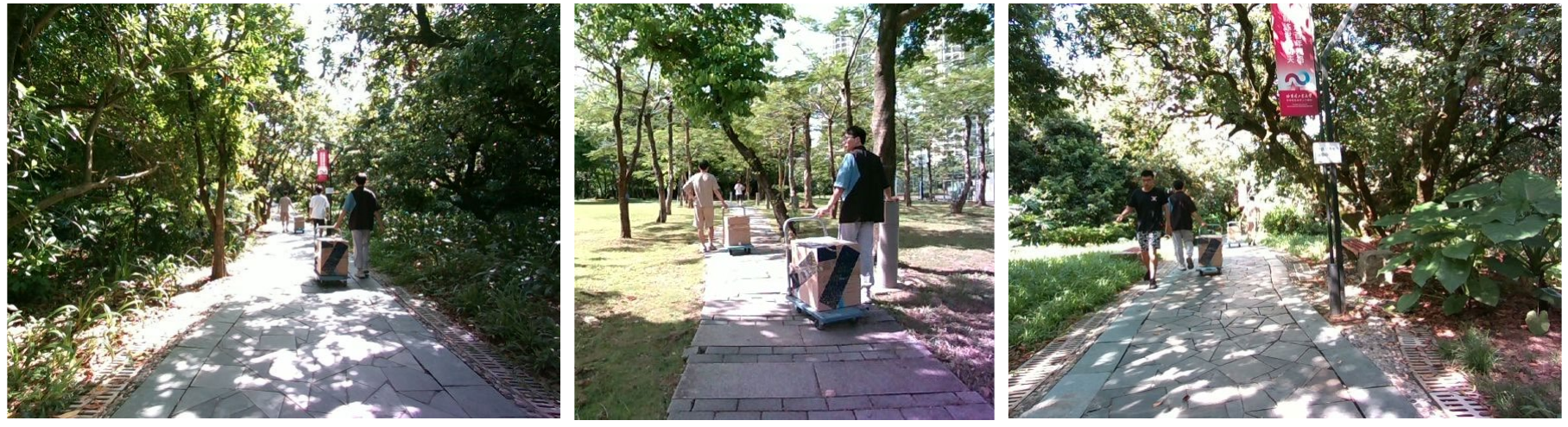}
	\caption{Real world environments. The targets, including two persons and two trolleys, move in the woods. The robot with sensors follows the targets.}
	\label{fig:training_environment_realworld}
\end{figure}

In addition, we also developed a physical robot platform, as shown in Fig. \ref{fig:the robot physical platform}. The mobile platform is a small off-road vehicle based on the Ackerman model. The sensors include RS-Ruby Lidar (128-line), RealSense D455 camera, and XSENS MTi-G-710 IMU. The physical experiments of a robot tracking moving target objects in the woods of Shenzhen University Town were carried out. The environments are shown in Fig. \ref{fig:training_environment_realworld}. The four pieces of data exp00-{exp03} were recorded. exp03 segment data is used as training dataset and others are used as testing datasets.

\subsection{Evaluation of 3D multi-object tracking}
We use three metrics, including Multi-Object Tracking Accuracy (MOTA), Multi-Object Tracking Precision (MOTP), and Object Class Accuracy (OCA), to evaluate the results of 3D multi-object tracking. The definitions are described as follows: 

(1) Multi-Object Tracking Accuracy (MOTA): 
The proportion of tracking instances that exclude missed detection, false detection, and wrong matching to the total number of real trajectories is used to evaluate the accuracy of multi-object tracking algorithms. 
A successful recall of a tracking instance means that the 3D Euclidean distance between the estimated position of tracked instance and the ground truth position is fewer than 0.5 m. 
The definition of missed detection is that the estimated tracking instances number of the current data frame is less than the true value. 
The definition of false detection is that the estimated tracking instances number of the current data frame is more than the true value. 
The definition of false match is that the estimated tracking instance ID of the current data frame is different from the previous corresponding ground truth ID. 
\begin{equation}
	MOTA=1 - \frac{\sum_i \left(FN_i + FP_i + ID_{sw}\right)}{\sum_i GT}
	\label{equ:MOTA}
\end{equation}
where $FN_i$, $FP_i$, $ID_{sw}$, and $GT$ represent the number of false negatives, false positives, false matches, and ground truth of i-th frame, respectively.

(2) Multi-Object Tracking Precision (MOTP): 
It evaluates the error between the estimated tracking instance position and the true value. 
\begin{equation}
	MOTP = \frac{\sum_{j,i} e_i^j}{\sum_i TP_i}
	\label{equ:MOTP}
\end{equation}
where $e_i^j$ represents the Euclidean distance error between the position of the j-th tracked instance estimated in i-th frame and the true value, and $TP_i$ represents the number of matches between the tracked instance in i-th frame and the ground truth.

(3) Object Class Accuracy (OCA): 
It evaluates the estimated class accuracy of tracking instances. 
\begin{equation}
	OCA = \frac{\sum_i TL_i}{\sum_i TP_i}
	\label{equ:OCA}
\end{equation}
where $TL_i$ represents the correct number of tracked instance categories estimated in i-th frame, and $TP_i$ represents the number of tracked instances in i-th frame that matches the true value.

The existing tracking algorithm-AB3DMOT \cite{XinshuoWeng2020AB3DMOTAB} was reproduced as a comparison method and tested on
the simulated datasets with the proposed tracking method. To ensure fairness in comparing the tracking module, the interface of the detection module in AB3DMOT was changed to be consistent with ours.
The statistical results are shown in TABLE \ref{tab:The experimental results of the tracking algorithm}.
Compared to AB3DMOT, the proposed tracking algorithm improves the metric of MOTA because the proposed adaptive threshold algorithm makes the matching more accurate and reduces the false negative and false positive detections.

\begin{table}[htbp]
	\caption{The experimental results of the tracking algorithm}
	\label{tab:The experimental results of the tracking algorithm}
    \centering
		\begin{tabular}{ccccc}
		\toprule[1.5pt] 
			dataset & method & MOTA(\%)   & MOTP(m)  & OCA(\%) \\
		\midrule
			\multirow{2}{*}{sim00}
			&AB3DMOT & 89.32 & 0.109 & 97.21 \\
			&Ours & \textbf{89.61} & 0.109 & \textbf{98.81} \\ \hline

			\multirow{2}{*}{sim01}
			&AB3DMOT & 78.78 & 0.112 & \textbf{98.33} \\
			&Ours & \textbf{80.23} & 0.112 & 97.85 \\ \hline

			\multirow{2}{*}{sim02}
			&AB3DMOT & 86.61 & 0.112 & 97.23 \\
			&Ours & \textbf{86.85} & 0.112 & \textbf{98.51} \\
		\bottomrule
		\end{tabular}
\end{table}

\subsection{Evaluation of 3D trajectory prediction}
Average displacement error (ADE) and final displacement error (FDE) are commonly used to evaluate trajectory prediction accuracy. ADE represents the average Euclidean distance difference between each predicted position and the ground truth position, calculated by \eqref{equ:ADE}:
\begin{equation}
	ADE =\frac{\sum_{i=1}^{m} \sum_{j=f_{obs}+1}^{f_{pre}} e_{ij} }
		{m\left(f_{pre} - f_{obs}\right)} 
	\label{equ:ADE}
\end{equation}
where $m$ represents the number of trajectories, $f_{obs}$ indicates the current observation frame, $f_{pre}$ is the last frame of the prediction time, $e_{ij}$ indicates the difference between the predicted position and the ground truth of the $j$th frame of the $i$th tracking instance Euclidean distance.

FDE represents the Euclidean distance between the predicted end position and the ground truth position, calculated according to \eqref{equ:FDE}:
\begin{equation}
	FDE =\frac{\sum_{i=1}^{m} e_{i,f_{pre}} }
		{m} 
	\label{equ:FDE}
\end{equation}

We use the least squares trajectory appropriate method (LSF) and Kalman filter (KF) for trajectory prediction as comparison. The results are shown in TABLE \ref{tab:Trajectory prediction experiment}.
ADE and FDE obtained by the proposed 3D trajectory prediction algorithm are smaller than the least square fitting and Kalman filter trajectory prediction algorithms.
\begin{table}[htbp]
	\caption{Trajectory prediction experiment}
	\label{tab:Trajectory prediction experiment}
    \centering
		\begin{tabular}{ccccc}
			\toprule[1.5pt] 
				dataset & method & ADE(m)   & FDE(m)   \\
			\midrule
				\multirow{3}{*}{sim00}
				&KF          & 2.157 & 5.010 \\
				&LSF & 3.235 & 7.853 \\
				&Ours & \textbf{2.065} & \textbf{4.626} \\ \hline
	
				\multirow{3}{*}{sim01}
				&KF          & 2.317 & 5.290  \\
				&LSF & 3.636 & 8.752  \\
				&Ours & \textbf{2.195} & \textbf{4.848} \\ \hline
	
				\multirow{3}{*}{sim02}
				&KF          & 2.342 & 5.357 \\
				&LSF & 3.403 & 8.243 \\
				&Ours & \textbf{2.247} & \textbf{4.998} \\
			\bottomrule
		\end{tabular}
\end{table}

\subsection{Evaluation of dynamic SLAM}
\begin{figure}[htpb]
	\centering
	\includegraphics[width = 3.4in]{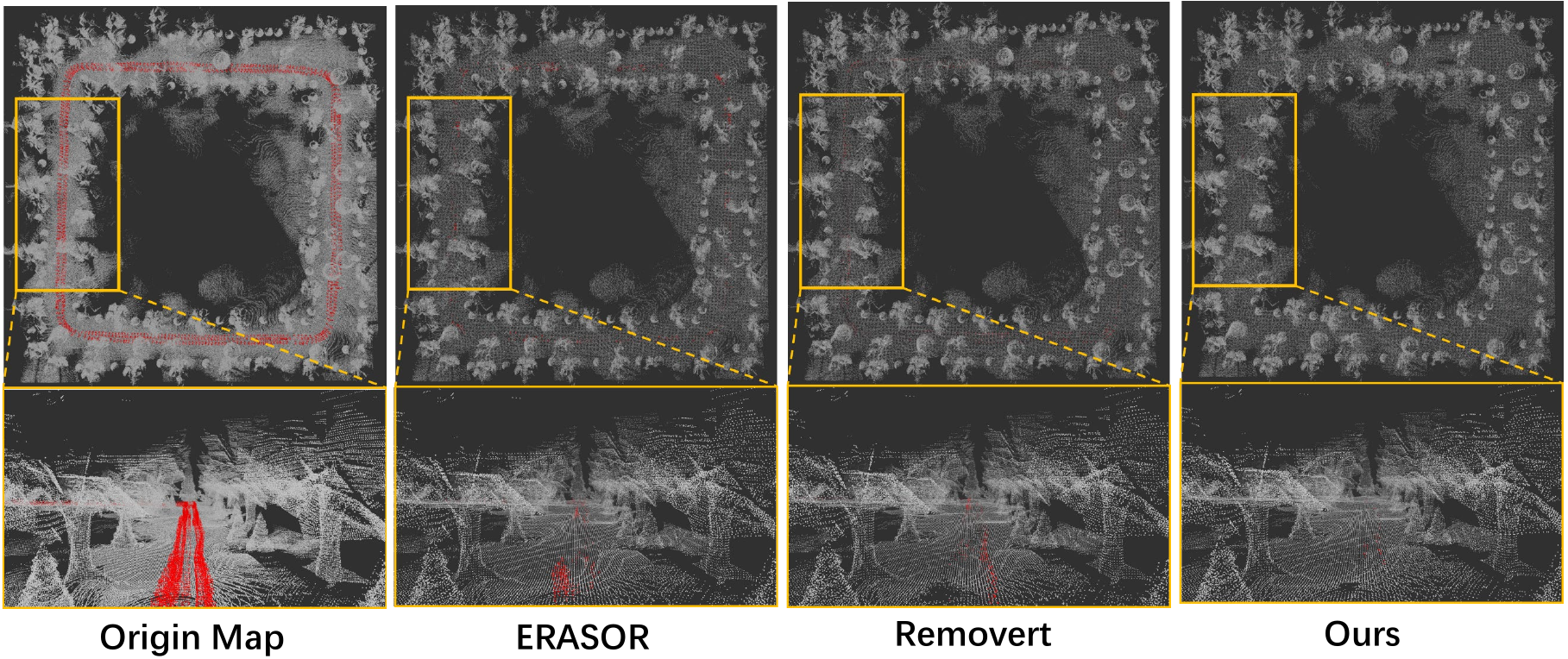}
	\caption{Comparison of mapping result for different dynamic SLAM methods. The red points denote the unwanted dynamic points. }
	\label{fig:dynamic_slam_result}
\end{figure}

We used ERASOR \cite{lim2021erasor} and Removert \cite{GiseopKim2020RemoveTR} as comparison methods to evaluate the performance of dynamic SLAM after removing dynamic points.
As shown in Fig. \ref{fig:dynamic_slam_result}, the original map in the left sub-figure has many dynamic points (blue oval mark). And our method removes dynamic points more thoroughly compared to ERASOR \cite{lim2021erasor} and Removert \cite{GiseopKim2020RemoveTR}.

To evaluate the quality of the static map saved after removing the dynamic points, we use the static point preservation rate (PR) and dynamic point rejection rate (RR) \cite{lim2021erasor} as the evaluation metric. The specific calculation formulas of PR and RR are shown in \eqref{equ:PR} and \eqref{equ:RR} as follows:
\begin{equation}
	PR = \frac{N_{sp}}{N_{sa}}
	\label{equ:PR}
\end{equation}
\begin{equation}
	RR = 1 - \frac{N_{dp}}{N_{da}}
	\label{equ:RR}
\end{equation}
where $N_{sp}$ represents the number of static points preserved in the static map; $N_{sa}$ represents the total number of static points in the original map; $N_{dp}$ represents the number of dynamic points kept in the static map; $N_{da}$ represents the total number of dynamic points in the original map. The number of static and dynamic points in the static map is obtained through grid statistics of the corresponding positions of the original map and static maps' corresponding positions. The grid size set here is 0.2 m.
The $F_1\ score$, calculated by PR and RR, is also used for evaluation and is defined as follows:
\begin{equation}
	F_1\ score = 2 \frac{PR*RR}{PR+RR}
	\label{equ:F_1_score}
\end{equation}

Three datasets ({sim00}-{sim02}) are used for testing. ERASOR, Removert, and the proposed method are used to build static maps. The results of the preservation rate (PR), rejection rate (RR), and $F_1\ score$ are shown in TABLE \ref{tab:The static map quality}. The static map obtained by the proposed method has the best results on the three test datasets.
\begin{table}[htbp]
	\caption{The static map quality}
	\label{tab:The static map quality}
    \centering
		\begin{tabular}{ccccc}
		\toprule[1.5pt] 
			dataset & method & PR(\%)   & RR(\%)  & $F_1\ score$ \\
		\midrule
			\multirow{3}{*}{sim00}
			&ERASOR & 98.05 & 90.05 & 0.938 \\
			&Removert & 98.03 & 88.97 & 0.932 \\
			&Ours & \textbf{99.03} & \textbf{99.70} & \textbf{0.993} \\ \hline

			\multirow{3}{*}{sim01}
			&ERASOR & 92.95 & 88.49 & 0.906 \\
			&Removert & 98.67 & 79.24 & 0.878 \\
			&Ours & \textbf{98.74} & \textbf{99.65} & \textbf{0.991} \\ \hline

			\multirow{3}{*}{sim02}
			&ERASOR & 96.55 & 88.91 & 0.925 \\
			&Removert & 98.68 & 88.05 & 0.930 \\
			&Ours & \textbf{99.16} & \textbf{99.69} & \textbf{0.994} \\
		\bottomrule
		\end{tabular}
\end{table}

\subsection{Evaluation of time-consuming}
\begin{figure}[htpb]
	\centering
	\includegraphics[width = 0.45\textwidth]{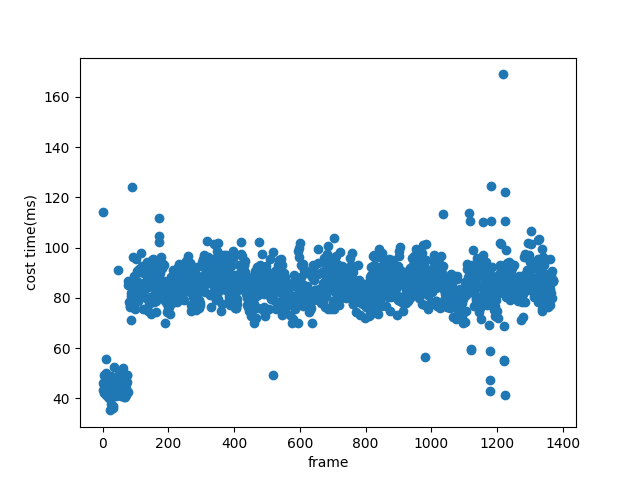}
	\caption{The time-consuming of the proposed 3D multi-object detection, tracking, and prediction system}
	\label{fig:The calculation time of the entire system}
\end{figure}

We also evaluate the time-consuming of the proposed system. The desktop hardware configuration is CPU: AMD Ryzen 9 3950x; GPU: NVIDIA GeForce RTX 3080.
The result is shown in Fig. \ref{fig:The calculation time of the entire system}. The average time-consuming of the entire system, including 3D multi-object detection, tracking, and prediction, is 83.27 ms. Therefore, the proposed system can run in real-time.

\section{CONCLUSION}
This paper proposes a real-time framework integrating 3D object detection, multi-object tracking, 3D trajectory prediction, and dynamic SLAM. The threshold-adaptive 3D multi-object tracking algorithm tracks multi-object well. The improved 3D trajectory prediction method that combines MANTRA and Kalman filter enhances the prediction accuracy and robustness. And the dynamic SLAM removes dynamic points based on the object detection according to the multi-object tracking results, making the removal more thorough. 
In the future, we will consider improving the robustness of the detection system when the robot travels in rough terrain and complex outdoor environments.

\normalem
\bibliographystyle{IEEEtran}
\bibliography{ref.bib}

\end{document}